\documentclass[conference]{IEEEtran}
\IEEEoverridecommandlockouts

\usepackage{cite}
\usepackage{amsmath,amssymb,amsfonts}
\usepackage{algorithmic}
\usepackage{graphicx}
\usepackage{textcomp}
\usepackage{xcolor}
\def\BibTeX{{\rm B\kern-.05em{\sc i\kern-.025em b}\kern-.08em
    T\kern-.1667em\lower.7ex\hbox{E}\kern-.125emX}}

\usepackage{booktabs}
\usepackage{tabularx}
\usepackage{multirow}
\usepackage{tikz}
\usepackage{url}
\usepackage{hyperref}
\usepackage[capitalise]{cleveref}

\graphicspath{{./figures}}
\DeclareGraphicsExtensions{.pdf,.jpeg,.jpg,.png}

\newcolumntype{V}{>{\raggedright\arraybackslash}X}
\newcolumntype{Y}{>{\centering\arraybackslash}X}
\newcolumntype{Z}{>{\raggedleft\arraybackslash}X}

\newcolumntype{v}{>{\hsize=.75\hsize}V}
\newcolumntype{y}{>{\hsize=.75\hsize}Y}
\newcolumntype{z}{>{\hsize=.75\hsize}Z}

\begin{document}

\title{Cross-Layer Cache Aggregation for Token Reduction in Ultra-Fine-Grained Image Recognition\\
\thanks{This work was supported by the National Science and Technology Council, Taiwan, under grant NSTC 111-2221-E-A49-092-MY3, NSTC 113-2640-E-A49-005, NSTC 112-2221-E-007-079-MY3 and NSTC 113-2218-E-007-020. We thank NYCU HPC for providing computational and storage resources.}
}

\author{
    Edwin Arkel Rios\textsuperscript{\dag}, Jansen Christopher Yuanda\textsuperscript{\ddag}, Vincent Leon Ghanz\textsuperscript{\ddag}, Cheng-Wei Yu\textsuperscript{\dag}, Bo-Cheng Lai\textsuperscript{\dag}, Min-Chun Hu\textsuperscript{\ddag}\\
    \textsuperscript{\dag}\textit{National Yang Ming Chiao Tung University, Taiwan}, \textsuperscript{\ddag}\textit{National Tsing Hua University, Taiwan}\\
}

\maketitle

\begin{tikzpicture}[remember picture,overlay]
\node at (current page.south west) [anchor=south west]
{\footnotesize \hspace{12mm} \begin{minipage}{0.99\textwidth}\begin{center}
\noindent © 20XX IEEE. Personal use of this material is permitted.  Permission from IEEE must be obtained for all other uses, in any current or future media, including reprinting/republishing this material for advertising or promotional purposes, creating new collective works, for resale or redistribution to servers or lists, or reuse of any copyrighted component of this work in other works..
\vspace{8mm}
\end{center}\end{minipage}};
\end{tikzpicture}

\begin{abstract}

    Ultra-fine-grained image recognition (UFGIR) is a challenging task that involves classifying images within a macro-category. While traditional FGIR deals with classifying different species, UFGIR goes beyond by classifying sub-categories within a species such as cultivars of a plant. In recent times the usage of Vision Transformer-based backbones has allowed methods to obtain outstanding recognition performances in this task but this comes at a significant cost in terms of computation specially since this task significantly benefits from incorporating higher resolution images. Therefore, techniques such as token reduction have emerged to reduce the computational cost. However, dropping tokens leads to loss of essential information for fine-grained categories, specially as the token keep rate is reduced. Therefore, to counteract the loss of information brought by the usage of token reduction we propose a novel Cross-Layer Aggregation Classification Head and a Cross-Layer Cache mechanism to recover and access information from previous layers in later locations. Extensive experiments covering more than 2000 runs across diverse settings including 5 datasets, 9 backbones, 7 token reduction methods, 5 keep rates, and 2 image sizes demonstrate the effectiveness of the proposed plug-and-play modules and allow us to push the boundaries of accuracy vs cost for UFGIR by reducing the kept tokens to extremely low ratios of up to 10\% while maintaining a competitive accuracy to state-of-the-art models. Code is available at: \url{https://github.com/arkel23/CLCA}



\end{abstract}


\section{Introduction}
\label{sec_intro}

Fine-grained image recognition (FGIR) involves classifying images into sub-categories within a larger macro-category \cite{wei_fine-grained_2021}. While conventional FGIR \cite{wei_fine-grained_2021} classifies objects usually up to species-level granularity, ultra-FGIR (UFGIR) may categorize classes at a finer level, such as cultivars of a plant. It has practical applications in fields such as agriculture \cite{yu_benchmark_2021}.

It is challenging due to small inter-class and large intra-class variations, but in recent years there has been a significant improvement in performance in terms of accuracy. Previous work has noted how this may be driven by backbone choices  \cite{ye_image_2024}. In particular, the Vision Transformer (ViT) \cite{dosovitskiy_image_2020} has brought a large performance improvement across tasks \cite{he_transfg_2022, wang_feature_2021}; the global receptive field of the ViT's attention mechanism provides it an advantage in handling long-range dependencies, allowing it to effectively discover and learn dependencies between discriminative features within an image \cite{ye_image_2024}. 

However, this accuracy improvement has come at a cost in terms of computational resources required for training and inference \cite{ye_image_2024}. ViT-based models have higher FLOPs compared to equivalent CNN-based models, specially as the image size is increased. This is due to the quadratic complexity $\mathcal{O}(N^2)$ of the self-attention mechanism with respect to the sequence length $N$. Specialized FGIR models employing ViT as a backbone \cite{he_transfg_2022} may even exhibit cubic complexity $\mathcal{O}(N^3)$ due to their usage of aggregated attention  \cite{abnar_quantifying_2020}.

Employing higher-resolution images in FGIR is common in order to identify the small differences between classes \cite{wang_efficient_2023, guo_fine-grained_2023}. In order to effectively make use of current and next-generation scaled models, specially in low-resource settings, it is crucial to increase the efficiency of such ViT-based models; token reduction is a promising direction to increase their efficiency.

Token Reduction (TR) proposes dropping \cite{rao_dynamicvit_2021, fayyaz_adaptive_2022} or combining \cite{renggli_learning_2022, bolya_token_2023} unimportant tokens from the sequence in order to reduce the number of tokens to be processed; it is also known as token or input pruning. In computer vision these tokens represent image patches. By reducing the number the tokens the FLOPs, latency, and memory consumption could be reduced, while throughput can be increased \cite{rao_dynamicvit_2021, tang_patch_2022}.

Different TR methods use different criteria for selecting or merging tokens; some use attention \cite{liang_evit_2021, fayyaz_adaptive_2022}, while others employ learnable modules \cite{rao_dynamicvit_2021, zong_self-slimmed_2022}. However, regardless of the TR selection criteria, the process of TR inevitably leads to the loss of valuable information for discrimination, specially as the keep ratio (KR) of tokens is reduced.

Therefore, to counteract the loss of information brought by the TR process we propose two novel mechanisms: 1) a Cross-Layer Aggregation (CLA) Classification Head which incorporates skip connections for intermediate features of a transformer directly into the classification module, and 2) a Cross-Layer Cache (CLC) structure to store and recover cross-layer information vital for fine-grained classification.

\begin{figure}[!htb]

    \vspace{-0.25cm}

    \begin{center} 
        \includegraphics[width=0.8\linewidth]{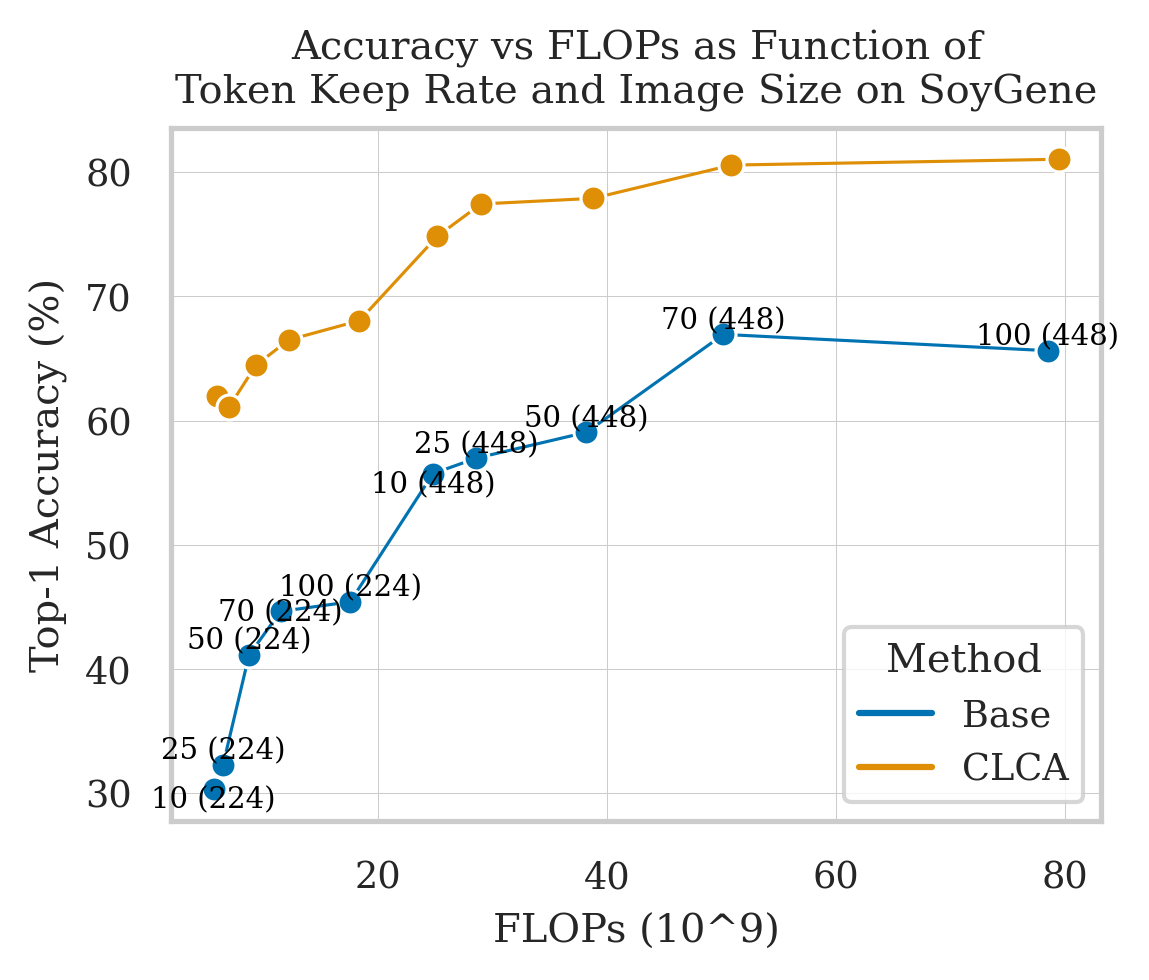}
    \end{center}

    \vspace{-0.25cm}

    \caption{
    Accuracy vs FLOPs for DeiT3 \cite{touvron_deit_2022} with EViT \cite{liang_evit_2021} token reduction method with (CLCA) and without (Base) our proposed modules on the SoyGene dataset \cite{yu_benchmark_2021} as a function of different keep rates (10, 25, 50, 70 and 100\%) and image sizes (in parenthesis).}

    \label{figure_accuracy_vs_flops}

    \vspace{-0.25cm}

\end{figure}

We refer to our full method method, incorporating both the CLA Head and the CLC, as Cross-Layer Cache Aggregation (CLCA). CLCA allows us to push the boundary of TR methods, by employing extremely low token keep ratios (KR) of up to 10\%, while keeping an accuracy that is on-par with (or even surpass in certain cases) state-of-the-art (SotA) methods in UFGIR. We demonstrate the performance improvement brought by our proposed contributions with different token KRs, and image sizes for the SoyGene dataset \cite{yu_benchmark_2021} in \cref{figure_accuracy_vs_flops}.

Our contributions are as follows:
\begin{enumerate}
    \item We propose a novel Cross-Layer Aggregation Classification (CLA) Head to expedite information from intermediate layers of a transformer directly into the classification module.

    \item We propose the Cross-Layer Cache (CLC) structure to counteract the loss of information brought by token reduction. The CLC allows us to access features from previous layer at deeper layers in order to facilitate fine-grained classification. 

    \item We conduct extensive experiments on 5 UFGIR datasets with a variety of TR schemes, token keep rates, and pretrained backbones. Across more than 2000 runs, the proposed model, CLCA, combining the CLA and CLC plug-and-play modules, significantly and consistently improves the accuracy across diverse settings, while incurring minimal increase in cost. Furthermore, it enables us to reduce the TR KR up to 10\%, drastically reducing computational costs, while maintaining an accuracy that is on par with models with much higher cost as measured by the number of floating-point-operations (FLOPs).
\end{enumerate}

\section{Related Work}
\label{sec_related}

\subsection{Ultra-Fine-Grained Image Recognition}
\label{ssec_fgir}

UFGIR methods make use of generic image recognition backbones and equip them with modules to select and aggregate discriminative features \cite{he_transfg_2022,wang_feature_2021} or employ loss functions and tasks \cite{fang_learning_2024} to guide models to more effectively make use of fine-grained features. 

However, many of these methods employ transformer-based backbones with a large computational cost, specially as the image size increases due to the quadratic complexity of the self-attention operator.

\subsection{Token Reduction}
\label{ssec_tr}

Token reduction methods can be broadly categorized into token pruning and token merging. Token pruning involves the dropping of a number of uninformative tokens at certain layers. For instance, DynamicViT \cite{rao_dynamicvit_2021} drops tokens with a static keep rate based on decisions predicted by a learnable MLP module, while ATS \cite{fayyaz_adaptive_2022} drops tokens with a dynamic keep rate based on attention scores. Conversely, token merging reduces the total number of tokens by combining multiple tokens into one. Zeng et al. \cite{zeng_not_2022} implement this via the DPC-KNN clustering algorithm, while Zong et al. \cite{zong_self-slimmed_2022} adopt a learnable weight approach, resulting in slimmed rather than dropped tokens.

\section{Proposed Method: CLCA}
\label{sec_method}

\begin{figure*}[!htb]
    
    \vspace{-0.5cm}

    \begin{center}
        \includegraphics[width=1.0\linewidth]{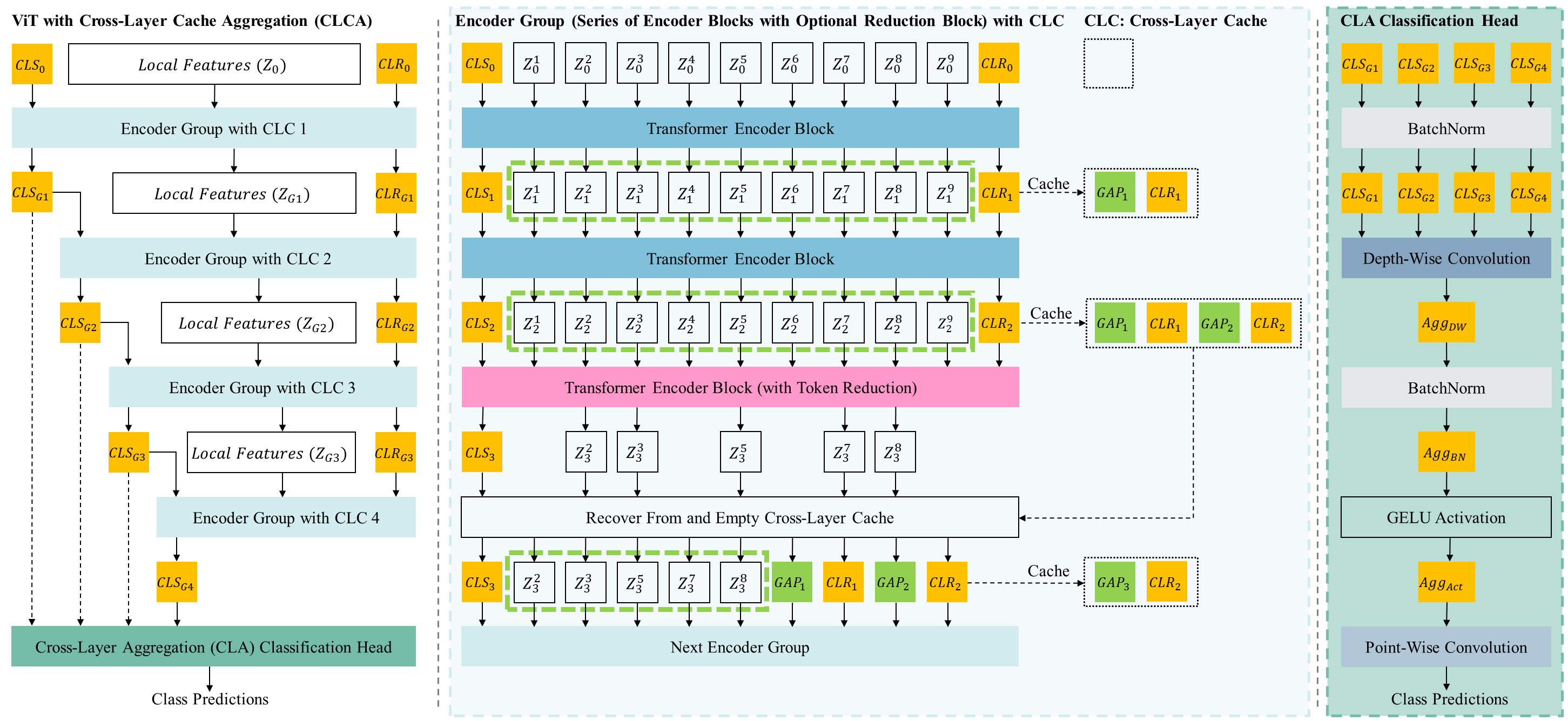}
    \end{center}

    \vspace{-0.5cm}
    
    \caption{From top to bottom, on the left is the overview for a ViT with the proposed Cross-Layer Cache Aggregation (CLCA). A learnable CLS and Cross-Layer Register (CLR) tokens are attached to a sequence of local features corresponding to image patches. This sequence is passed through a series of ViT Encoder Groups equipped with the proposed Cross-Layer Cache (CLC). Each Encoder Group is composed of a series of transformer encoder blocks followed by a reduction block that drops tokens at the end.  After each encoder block we cache the spatial Global Average Pooling (GAP) of the local features, the tokens surrounded by the green dashed box, and the last token in the sequence (the CLR). After the reduction block we recover intermediate features from previous layers from the CLC, empty it, and then cache the current layer outputs. We repeat the process for each encoder group. Thereafter, the CLS token outputs of each Encoder Group are processed by the Cross-Layer Aggregation (CLA) Classification Head. These CLS tokens are first normalized using a BatchNorm layer, then forwarded through a Depth-Wise Convolution layer that combines cross-layer features for each channel separately. A second BatchNorm and activation function are employed for learning non-linear interactions between intermediate features. Finally, the vector of features is passed through a Point-Wise Convolution that outputs the classification predictions.}

    \label{figure_overview_clca}

    \vspace{-0.5cm}

\end{figure*}

The overview of our proposed method is shown in \cref{figure_overview_clca}. Our method is based on a generic Vision Transformer with token reduction. The transformer encoder blocks are divided into four groups based on the token reduction locations and the last layer. Features in each group are processed using a series of encoder blocks followed by a token reduction and cache recovery operations. We aggregate intermediate features corresponding to the outputs of the CLS token of each group. These features are first normalized, then cross-layer channel-wise information is aggregated through a depth-wise convolution. We incorporate a non-linear activation function to model complex relations between features. Finally, a point-wise convolution projects the aggregated features into a vector of classification predictions. 

\subsection{Vision Transformer Encoder Group with Token Reduction}
\label{ssec_vit}

Images are patchified using a convolution with kernel size $P$ and flattened into a 1D sequence of $D$ channels with length $N = (S_1 / P) \times (S_2 / P)$, where $S_1$ and $S_2$ represent the image width and height. A learnable CLS token is appended at the start of the sequence. Learnable positional embeddings are added to incorporate spatial information. 

This sequence is then processed through a series of $L$ transformer encoder blocks, divided into $g$ groups, each consisting of multi-head self-attention (MHSA) and position-wise feed-forward networks (PWFFN).

At the last encoder block of each group we reduce tokens according to each TR scheme. Our method can be combined with most token reduction methodologies but we focus on EViT \cite{liang_evit_2021}. EViT selects tokens to keep based on the average attention-score of the CLS token with respect to the other tokens as computed in the MHSA block. 

Specifically, the top-$k$ tokens with highest attention based on the CLS token attention $\mathbf{A}_0$ are kept and the remaining tokens are merged using a weighted average operation. Then the kept tokens and the fused token will be concatenated and forwarded to the PWFFN block as normally.

\subsection{Cross-Layer Aggregation (CLA) Head}
\label{ssec_cla}

To prevent information loss we fast-forward intermediate features of the network directly into the classification layer. Specifically, we forward the CLS token output of each encoder group (after each reduction location and the last layer) into the proposed CLA Head, shown in the right block of \cref{figure_overview_clca}.

In the CLA Head we concatenate and reshape the tokens into a sequence of size $g$, denoted as $\mathbf{CLS} \in \mathbb{R}^{D \times g}$, before normalizing the features using a BatchNorm layer. Then, a Depth-Wise Convolution is employed to aggregate channel-wise interactions across layers. 



\vspace{-0.25cm}

\begin{equation}
    Agg = \text{DWConv}(\text{BN}([\text{CLS}_{G1}; \text{CLS}_{G2}; \ldots; \text{CLS}_{Gg}]))    
    \label{eq_cla_dwconv}
\end{equation}

The output is an aggregated token $Agg \in \mathbb{R}^{D\cdot \text{DWG}}$ where $\text{DWG}=2$ is the number of groups dedicated to each channel in the depth-wise convolution. This aggregated token is passed through a BatchNorm and GELU activation function in order to model complex non-linear relationships between features. Finally, a Point-Wise Convolution projects the aggregated features into a vector of class probabilities used for classification.

\begin{equation}
    preds = \text{PWConv}(\text{GELU}(\text{BN}(Agg)))
    \label{eq_cla_head_pwconv}
\end{equation}

\subsection{Cross-Layer Cache (CLC)}
\label{ssec_clc_overview}

Regardless of the TR scheme, the information encoded in the original tokens is lost. Since reduction is applied from early layers that process low-level features, it is unavoidable to lose information that deeper layers may require to make a distinction between fine-grained categories. Therefore, inspired by FFVT, \cite{wang_feature_2021} which aggregated low-level and middle-level features from the network in the last encoder layer of a ViT for improved classification performance, we incorporate a Cross-Layer Cache (CLC) structure to store local and discriminative features from intermediate layers for future usage. An overview of the CLC is shown in the middle block of \cref{figure_overview_clca}.

Specifically, after each transformer encoder block in an encoder group, we store the Global-Average-Pooling (GAP) pooled local features of the model. Furthermore, inspired by previous work which suggested that transformers need registers \cite {darcet_vision_2023} we append a learnable Cross-Layer Register (CLR) at the end of the sequence. After each encoder block we also store this CLR that aggregates discriminative information across layers for future use. 

After the reduction location or before the last layer, we access the CLC to recover the lost information tokens by appending the tokens stored in it to the original sequence. This allows our model to counteract the information loss that comes with token reduction and make usage of relevant cross-layer information for discrimination of fine-grained categories. Finally, the CLC is emptied so as to not repeat tokens in future recovery operations.

\section{Experiment Methodology}
\label{sec_experiments}

\begin{table}[!htb]

    \vspace{-0.25cm}
    
    \centering

    \caption{Top-1 accuracies (\%) and FLOPs for SotA methods in UFGIR. For CLCA we include the token keep rate in parenthesis.}


    \footnotesize{
    \begin{tabularx}{\linewidth}{X z z z Z}
        \toprule
        Method & Cotton & SoyAgeing & SoyGlobal & FLOPs ($10^9$) \\
        \midrule
        ViT \cite{dosovitskiy_image_2020} & 52.5 & 67.0 & 40.6 & 78.5 \\
        DeiT \cite{touvron_deit_2022} & 54.2 & 69.5 & 45.3 & 78.5 \\
        TransFG \cite{he_transfg_2022} & 54.6 & 72.2 & 21.2 & 447.9 \\
        SIM-Tr \cite{sun_sim-trans_2022} & 54.6 & 34.8 & \textbf{70.7} & 81.8 \\
        CSDNet \cite{fang_learning_2024} & \underline{57.9} & 75.4 & 56.3 & 78.5 \\
        CLCA (10\%) & 55.6 & \underline{87.4} & \underline{61.1} & \textbf{25.2} \\
        CLCA (70\%) & \textbf{67.8} & \textbf{88.3} & 58.2 & \underline{50.9} \\
        \bottomrule
    \end{tabularx}
    }

    \label{table_sota_soy}


\end{table}

Our experiments use 5 ultra-fine-grained leaves datasets collected by Yu et al. \cite{yu_benchmark_2021} where each category represents a confirmed cultivar name attached to the seed obtained from the genetic resource bank. Top-1 accuracies are averaged across three seeds \cite{gwilliam_fair_2021}. When applicable, the best values are highlighted in \textbf{bold} and the second best are \underline{underlined}.


We employ AdamW optimizer, 50 epochs, weight decay of 0.05, batch size of 32, two image sizes (224, 448), nine backbones (ViT \cite{dosovitskiy_image_2020}, DeiT \cite{touvron_training_2021}, DeiT3 IN1k \& IN21k \cite{touvron_deit_2022}, MIIL \cite{ridnik_imagenet-21k_2021}, MoCov3 \cite{chen_empirical_2021}, DINO \cite{caron_emerging_2021}, MAE \cite{he_masked_2022}, CLIP LAION \cite{schuhmann_laion-5b_2022}) five different keep rates (100, 70, 50, 25, and 10\%). The token reduction operations are done on the 4th, 7th, and 10th of a ViT B-16 model with 12 layers, therefore we recover tokens from the CLC after these same locations and also after layer 11. 


The proposed CLCA is agnostic to the TR method. We mainly apply it with EViT \cite{liang_evit_2021}, but we also apply it to methods from all four different TR paradigms: DynamicViT \cite{rao_dynamicvit_2021} for static pruning, ATS for dynamic pruning \cite{fayyaz_adaptive_2022}, SiT \cite{zong_self-slimmed_2022} and PatchMerger \cite{renggli_learning_2022} for soft merging, and DPCKNN \cite{zeng_not_2022} and ToMe \cite{bolya_token_2023} for hard merging.



\section{Results and Discussion}
\label{sec_results}




\subsection{Comparison of CLCA with Different TR Schemes}
\label{ssec_discussion_clca}

\begin{figure}[!htb]

    \vspace{-0.25cm}

    \centering
        \includegraphics[width=1.0\linewidth]{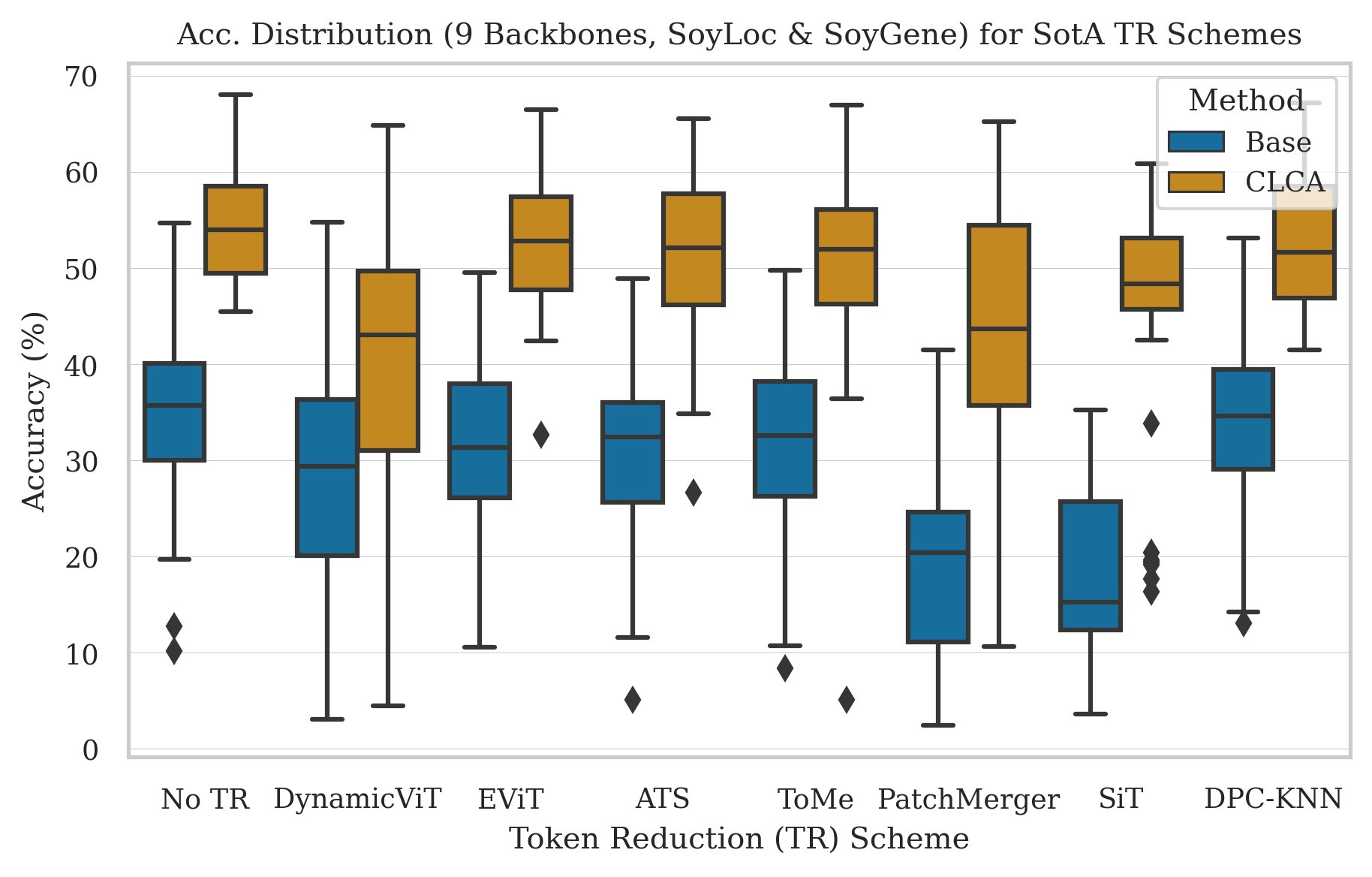}
    
    \vspace{-0.25cm}

    \caption{Distribution of accuracies for multiple keep rates (25, 50, and 70\%) on 9 different pretrained backbones as a function of the TR selection scheme (and the baseline without TR) on SoyLocal and SoyGene datasets with and without proposed CLCA.}

    \label{figure_box_acc_vs_tr}


\end{figure}

We plot the accuracy vs FLOPs as a function of image size (IS) and keep rate (KR) for the SoyGene dataset with EViT in \cref{figure_accuracy_vs_flops}. By incorporating CLCA the lower bound of the accuracy is raised to 61\% and to a maximum of 81\% compared to the original design which ranged from 30\% to 67\%. Models with CLCA are more robust to the TR KR and the IS compared to the original while incurring a negligible increase in cost as measured by the FLOPs.

These results are further validated by plotting the accuracies distribution (across 9 backbones, 3 keep rates, 3 random seeds, on 2 datasets) for different TR schemes with and without CLCA, as shown in \cref{figure_box_acc_vs_tr}. Our method consistently improves the performance across a wide variety of settings.

\subsection{Comparison with State-of-the-Art in UFGIR}
\label{ssec_discusion_sota}

We compare our method against state-of-the-art (SotA) methods using ViTs in the Cotton, SoyAgeing, and SoyGlobal datasets in \cref{table_sota_soy}. When compared to these we observe that our method obtains superior performance in terms of accuracy in the Cotton and SoyAgeing datasets. For the SoyGlobal dataset our method obtains the second best accuracy but the computational cost in terms of FLOPs is much lower.

\subsection{Discussion on Why CLCA Works}
\label{ssec_discusion_why_works}

\begin{figure}[!htb]

    \vspace{-0.25cm}

    \centering
        \includegraphics[width=0.6\linewidth]{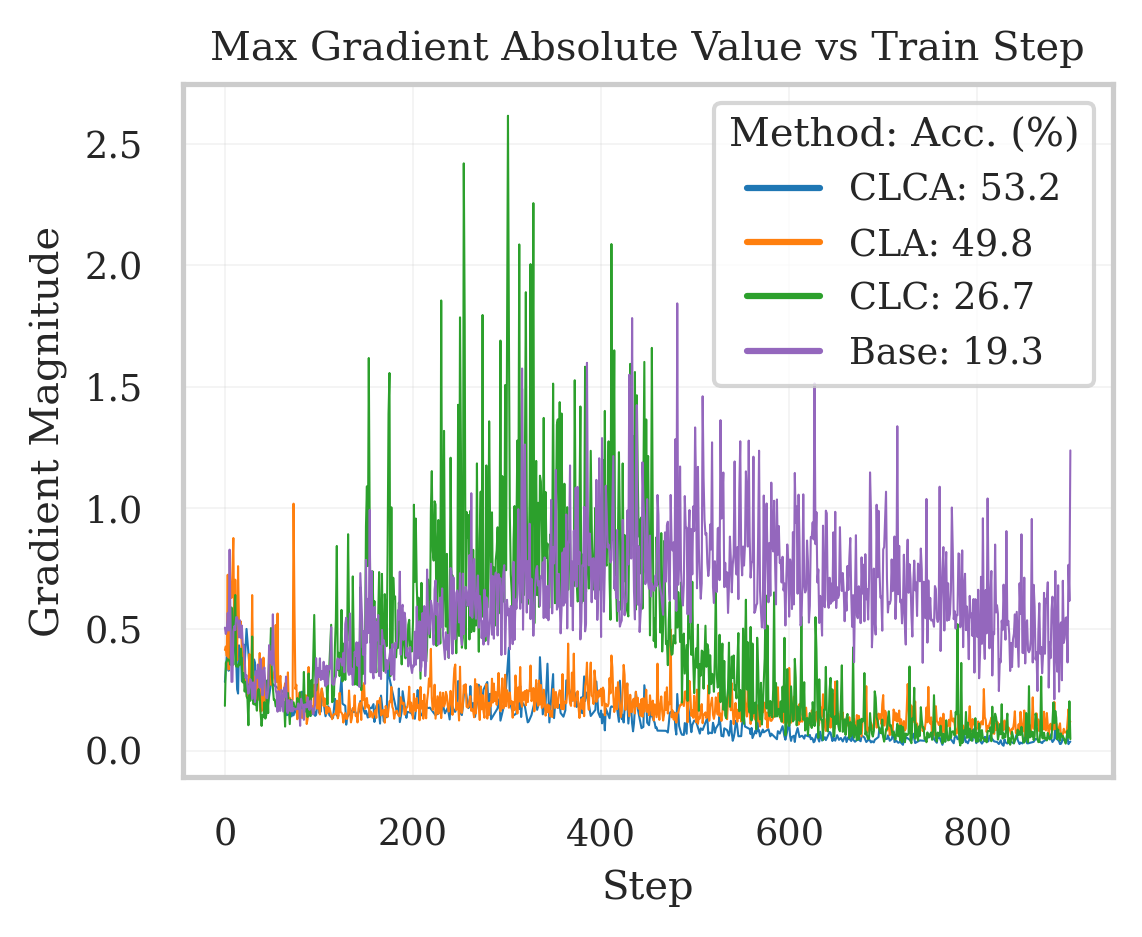}

    \vspace{-0.25cm}

    \caption{Ablation showing max magnitude of gradient across all layers of the model during training for DeiT3 B-16 \cite {touvron_deit_2022} on SoyLocal dataset with our proposed modules and the corresponding top-1 accuracy.}

    \label{figure_gradients}

\end{figure}

In \cref{figure_gradients} we plot the gradients during training for our model. Our analysis reveals that, prior to incorporating the proposed components, the gradients exhibit significant instability.

Therefore, inspired by previous which suggests that skip connections can smooth out the optimization landscape \cite{li_visualizing_2018} we incorporate information transfer in between 1) intermediate layers of the network to the (CLA) classification head, and 2) intermediate layers of an encoder group and the subsequent block through the usage of the Cross-Layer Cache (CLC). These proposed modules function as skip connections which have been shown to ease the training by avoiding spurious local optimum \cite{liu_towards_2019} and encouraging feature reuse \cite{huang_densely_2017}.

Furthermore, in UFGIR the usage of low-, middle-, and high-level features has been shown to be beneficial to distinguish between classes \cite{wang_feature_2021}. Our proposed modules provide shorter connections to these features, which has been shown to provide implicit deep supervision \cite{veit_residual_2016}.

\section{Conclusion}
\label{sec_conclusion}

In this paper we propose Cross-Layer Cache Aggregation (CLCA), a method to improve the performance of token reduction in ultra-fine-grained image recognition tasks. The proposed method integrates intermediate features from the network through a Cross-Layer Aggregation (CLA) classification head for robust predictions, and Cross-Layer Cache (CLC) to recover and integrate lost information from previous layers that is lost under the token reduction paradigm. The proposed methodology can be combined with a variety of mechanisms to reduce tokens, massively reducing the computational cost while consistently boosting the performance across different UFGIR datasets under a variety of settings.


{\small
\bibliographystyle{IEEEtran}
\bibliography{main}

\begin{thebibliography}{10}
\providecommand{\url}[1]{#1}
\csname url@samestyle\endcsname
\providecommand{\newblock}{\relax}
\providecommand{\bibinfo}[2]{#2}
\providecommand{\BIBentrySTDinterwordspacing}{\spaceskip=0pt\relax}
\providecommand{\BIBentryALTinterwordstretchfactor}{4}
\providecommand{\BIBentryALTinterwordspacing}{\spaceskip=\fontdimen2\font plus
\BIBentryALTinterwordstretchfactor\fontdimen3\font minus \fontdimen4\font\relax}
\providecommand{\BIBforeignlanguage}[2]{{%
\expandafter\ifx\csname l@#1\endcsname\relax
\typeout{** WARNING: IEEEtran.bst: No hyphenation pattern has been}%
\typeout{** loaded for the language `#1'. Using the pattern for}%
\typeout{** the default language instead.}%
\else
\language=\csname l@#1\endcsname
\fi
#2}}
\providecommand{\BIBdecl}{\relax}
\BIBdecl

\bibitem{wei_fine-grained_2021}
X.-S. Wei, Y.-Z. Song, O.~Mac~Aodha, J.~Wu, Y.~Peng, J.~Tang, J.~Yang, and S.~Belongie, ``Fine-{Grained} {Image} {Analysis} with {Deep} {Learning}: {A} {Survey},'' \emph{IEEE Transactions on Pattern Analysis and Machine Intelligence}, pp. 1--1, 2021, conference Name: IEEE Transactions on Pattern Analysis and Machine Intelligence.

\bibitem{yu_benchmark_2021}
X.~Yu, Y.~Zhao, Y.~Gao, X.~Yuan, and S.~Xiong, ``Benchmark {Platform} for {Ultra}-{Fine}-{Grained} {Visual} {Categorization} {Beyond} {Human} {Performance},'' in \emph{2021 {IEEE}/{CVF} {International} {Conference} on {Computer} {Vision} ({ICCV})}, Oct. 2021, pp. 10\,265--10\,275, iSSN: 2380-7504.

\bibitem{ye_image_2024}
S.~Ye, Y.~Wang, Q.~Peng, X.~You, and C.~L.~P. Chen, ``The {Image} {Data} and {Backbone} in {Weakly} {Supervised} {Fine}-{Grained} {Visual} {Categorization}: {A} {Revisit} and {Further} {Thinking},'' \emph{IEEE Transactions on Circuits and Systems for Video Technology}, vol.~34, no.~1, pp. 2--16, Jan. 2024, conference Name: IEEE Transactions on Circuits and Systems for Video Technology.

\bibitem{dosovitskiy_image_2020}
A.~Dosovitskiy, L.~Beyer, A.~Kolesnikov, D.~Weissenborn, X.~Zhai, T.~Unterthiner, M.~Dehghani, M.~Minderer, G.~Heigold, S.~Gelly, J.~Uszkoreit, and N.~Houlsby, ``An {Image} is {Worth} 16x16 {Words}: {Transformers} for {Image} {Recognition} at {Scale},'' \emph{arXiv:2010.11929 [cs]}, Oct. 2020, arXiv: 2010.11929.

\bibitem{he_transfg_2022}
J.~He, J.-N. Chen, S.~Liu, A.~Kortylewski, C.~Yang, Y.~Bai, and C.~Wang, ``\BIBforeignlanguage{en}{{TransFG}: {A} {Transformer} {Architecture} for {Fine}-{Grained} {Recognition}},'' in \emph{\BIBforeignlanguage{en}{Proceedings of the {First} {MiniCon} {Conference}}}, Feb. 2022.

\bibitem{wang_feature_2021}
J.~Wang, X.~Yu, and Y.~Gao, ``Feature {Fusion} {Vision} {Transformer} for {Fine}-{Grained} {Visual} {Categorization},'' in \emph{British {Machine} {Vision} {Conference} ({BMVC})}, Jul. 2021, arXiv: 2107.02341.

\bibitem{abnar_quantifying_2020}
S.~Abnar and W.~Zuidema, ``Quantifying {Attention} {Flow} in {Transformers},'' May 2020, arXiv:2005.00928 [cs].

\bibitem{wang_efficient_2023}
L.~Wang, J.~Zhang, J.~Tian, J.~Li, L.~Zhuo, and Q.~Tian, ``Efficient {Fine}-{Grained} {Object} {Recognition} in {High}-{Resolution} {Remote} {Sensing} {Images} {From} {Knowledge} {Distillation} to {Filter} {Grafting},'' \emph{IEEE Transactions on Geoscience and Remote Sensing}, vol.~61, pp. 1--16, 2023, conference Name: IEEE Transactions on Geoscience and Remote Sensing.

\bibitem{guo_fine-grained_2023}
B.~Guo, R.~Zhang, H.~Guo, W.~Yang, H.~Yu, P.~Zhang, and T.~Zou, ``Fine-{Grained} {Ship} {Detection} in {High}-{Resolution} {Satellite} {Images} {With} {Shape}-{Aware} {Feature} {Learning},'' \emph{IEEE Journal of Selected Topics in Applied Earth Observations and Remote Sensing}, vol.~16, pp. 1914--1926, 2023, conference Name: IEEE Journal of Selected Topics in Applied Earth Observations and Remote Sensing.

\bibitem{rao_dynamicvit_2021}
Y.~Rao, W.~Zhao, B.~Liu, J.~Lu, J.~Zhou, and C.-J. Hsieh, ``{DynamicViT}: {Efficient} {Vision} {Transformers} with {Dynamic} {Token} {Sparsification},'' Oct. 2021.

\bibitem{fayyaz_adaptive_2022}
M.~Fayyaz, S.~A. Koohpayegani, F.~R. Jafari, S.~Sengupta, H.~R.~V. Joze, E.~Sommerlade, H.~Pirsiavash, and J.~Gall, ``Adaptive {Token} {Sampling} {For} {Efficient} {Vision} {Transformers},'' Jul. 2022, arXiv:2111.15667 [cs].

\bibitem{renggli_learning_2022}
C.~Renggli, A.~S. Pinto, N.~Houlsby, B.~Mustafa, J.~Puigcerver, and C.~Riquelme, ``Learning to {Merge} {Tokens} in {Vision} {Transformers},'' Feb. 2022, arXiv:2202.12015 [cs].

\bibitem{bolya_token_2023}
D.~Bolya, C.-Y. Fu, X.~Dai, P.~Zhang, C.~Feichtenhofer, and J.~Hoffman, ``Token {Merging}: {Your} {ViT} {But} {Faster},'' Mar. 2023, arXiv:2210.09461 [cs].

\bibitem{tang_patch_2022}
Y.~Tang, K.~Han, Y.~Wang, C.~Xu, J.~Guo, C.~Xu, and D.~Tao, ``Patch {Slimming} for {Efficient} {Vision} {Transformers},'' Apr. 2022.

\bibitem{liang_evit_2021}
Y.~Liang, C.~Ge, Z.~Tong, Y.~Song, J.~Wang, and P.~Xie, ``\BIBforeignlanguage{en}{{EViT}: {Expediting} {Vision} {Transformers} via {Token} {Reorganizations}},'' Oct. 2021.

\bibitem{zong_self-slimmed_2022}
Z.~Zong, K.~Li, G.~Song, Y.~Wang, Y.~Qiao, B.~Leng, and Y.~Liu, ``Self-slimmed {Vision} {Transformer},'' Sep. 2022.

\bibitem{touvron_deit_2022}
H.~Touvron, M.~Cord, and H.~Jégou, ``{DeiT} {III}: {Revenge} of the {ViT},'' Apr. 2022, arXiv:2204.07118 [cs].

\bibitem{fang_learning_2024}
Z.~Fang, X.~Jiang, H.~Tang, and Z.~Li, ``Learning {Contrastive} {Self}-{Distillation} for {Ultra}-{Fine}-{Grained} {Visual} {Categorization} {Targeting} {Limited} {Samples},'' \emph{IEEE Transactions on Circuits and Systems for Video Technology}, pp. 1--1, 2024, conference Name: IEEE Transactions on Circuits and Systems for Video Technology.

\bibitem{zeng_not_2022}
W.~Zeng, S.~Jin, W.~Liu, C.~Qian, P.~Luo, W.~Ouyang, and X.~Wang, ``\BIBforeignlanguage{en}{Not {All} {Tokens} {Are} {Equal}: {Human}-centric {Visual} {Analysis} via {Token} {Clustering} {Transformer}},'' in \emph{\BIBforeignlanguage{en}{2022 {IEEE}/{CVF} {Conference} on {Computer} {Vision} and {Pattern} {Recognition} ({CVPR})}}.\hskip 1em plus 0.5em minus 0.4em\relax New Orleans, LA, USA: IEEE, Jun. 2022, pp. 11\,091--11\,101.

\bibitem{darcet_vision_2023}
T.~Darcet, M.~Oquab, J.~Mairal, and P.~Bojanowski, ``\BIBforeignlanguage{en}{Vision {Transformers} {Need} {Registers}},'' Oct. 2023.

\bibitem{sun_sim-trans_2022}
H.~Sun, X.~He, and Y.~Peng, ``{SIM}-{Trans}: {Structure} {Information} {Modeling} {Transformer} for {Fine}-grained {Visual} {Categorization},'' in \emph{Proceedings of the 30th {ACM} {International} {Conference} on {Multimedia}}, ser. {MM} '22.\hskip 1em plus 0.5em minus 0.4em\relax New York, NY, USA: Association for Computing Machinery, Oct. 2022, pp. 5853--5861.

\bibitem{gwilliam_fair_2021}
M.~Gwilliam, A.~Teuscher, C.~Anderson, and R.~Farrell, ``\BIBforeignlanguage{en}{Fair {Comparison}: {Quantifying} {Variance} in {Results} for {Fine}-{Grained} {Visual} {Categorization}},'' 2021, pp. 3309--3318.

\bibitem{touvron_training_2021}
H.~Touvron, M.~Cord, M.~Douze, F.~Massa, A.~Sablayrolles, and H.~Jégou, ``Training data-efficient image transformers \& distillation through attention,'' \emph{arXiv:2012.12877 [cs]}, Jan. 2021, arXiv: 2012.12877.

\bibitem{ridnik_imagenet-21k_2021}
T.~Ridnik, E.~Ben-Baruch, A.~Noy, and L.~Zelnik, ``\BIBforeignlanguage{en}{{ImageNet}-{21K} {Pretraining} for the {Masses}},'' \emph{\BIBforeignlanguage{en}{Proceedings of the Neural Information Processing Systems Track on Datasets and Benchmarks}}, vol.~1, Dec. 2021.

\bibitem{chen_empirical_2021}
X.~Chen, S.~Xie, and K.~He, ``\BIBforeignlanguage{en}{An {Empirical} {Study} of {Training} {Self}-{Supervised} {Vision} {Transformers}},'' in \emph{\BIBforeignlanguage{en}{2021 {IEEE}/{CVF} {International} {Conference} on {Computer} {Vision} ({ICCV})}}.\hskip 1em plus 0.5em minus 0.4em\relax Montreal, QC, Canada: IEEE, Oct. 2021, pp. 9620--9629.

\bibitem{caron_emerging_2021}
M.~Caron, H.~Touvron, I.~Misra, H.~Jegou, J.~Mairal, P.~Bojanowski, and A.~Joulin, ``Emerging {Properties} in {Self}-{Supervised} {Vision} {Transformers},'' in \emph{2021 {IEEE}/{CVF} {International} {Conference} on {Computer} {Vision} ({ICCV})}, Oct. 2021, pp. 9630--9640, iSSN: 2380-7504.

\bibitem{he_masked_2022}
K.~He, X.~Chen, S.~Xie, Y.~Li, P.~Dollar, and R.~Girshick, ``\BIBforeignlanguage{en}{Masked {Autoencoders} {Are} {Scalable} {Vision} {Learners}},'' in \emph{\BIBforeignlanguage{en}{2022 {IEEE}/{CVF} {Conference} on {Computer} {Vision} and {Pattern} {Recognition} ({CVPR})}}.\hskip 1em plus 0.5em minus 0.4em\relax New Orleans, LA, USA: IEEE, Jun. 2022, pp. 15\,979--15\,988.

\bibitem{schuhmann_laion-5b_2022}
C.~Schuhmann, R.~Beaumont, R.~Vencu, C.~Gordon, R.~Wightman, M.~Cherti, T.~Coombes, A.~Katta, C.~Mullis, M.~Wortsman, P.~Schramowski, S.~Kundurthy, K.~Crowson, L.~Schmidt, R.~Kaczmarczyk, and J.~Jitsev, ``\BIBforeignlanguage{en}{{LAION}-{5B}: {An} open large-scale dataset for training next generation image-text models},'' \emph{\BIBforeignlanguage{en}{Advances in Neural Information Processing Systems}}, vol.~35, pp. 25\,278--25\,294, Dec. 2022.

\bibitem{li_visualizing_2018}
H.~Li, Z.~Xu, G.~Taylor, C.~Studer, and T.~Goldstein, ``Visualizing the {Loss} {Landscape} of {Neural} {Nets},'' in \emph{Advances in {Neural} {Information} {Processing} {Systems}}, vol.~31.\hskip 1em plus 0.5em minus 0.4em\relax Curran Associates, Inc., 2018.

\bibitem{liu_towards_2019}
T.~Liu, M.~Chen, M.~Zhou, S.~S. Du, E.~Zhou, and T.~Zhao, ``Towards {Understanding} the {Importance} of {Shortcut} {Connections} in {Residual} {Networks},'' in \emph{Advances in {Neural} {Information} {Processing} {Systems}}, vol.~32.\hskip 1em plus 0.5em minus 0.4em\relax Curran Associates, Inc., 2019.

\bibitem{huang_densely_2017}
G.~Huang, Z.~Liu, L.~Van Der~Maaten, and K.~Q. Weinberger, ``Densely {Connected} {Convolutional} {Networks},'' in \emph{2017 {IEEE} {Conference} on {Computer} {Vision} and {Pattern} {Recognition} ({CVPR})}, Jul. 2017, pp. 2261--2269, iSSN: 1063-6919.

\bibitem{veit_residual_2016}
A.~Veit, M.~Wilber, and S.~Belongie, ``Residual networks behave like ensembles of relatively shallow networks,'' in \emph{Proceedings of the 30th {International} {Conference} on {Neural} {Information} {Processing} {Systems}}, ser. {NIPS}'16.\hskip 1em plus 0.5em minus 0.4em\relax Red Hook, NY, USA: Curran Associates Inc., Dec. 2016, pp. 550--558.

\end{thebibliography}
}

\end{document}